\documentclass[11pt,a4paper]{article}
\usepackage[T1]{fontenc}
\usepackage[utf8]{inputenc}
\usepackage{lmodern}
\usepackage[margin=25mm]{geometry}
\usepackage{microtype}
\usepackage{amsmath,amssymb}
\usepackage{booktabs,tabularx,array}
\usepackage{listings}
\usepackage{placeins}
\usepackage{xurl}
\usepackage[hidelinks]{hyperref}
\newcolumntype{Y}{>{\raggedright\arraybackslash}X}
\hypersetup{pdftitle={DKnown-Jev: Fine-Tuning and Evaluation of a JEV-Style Model for Tasks Involving Normative Knowledge},pdfauthor={Qi Li, Jianjun Xu, Pingtao Wei}}
\title{DKnown-Jev: Fine-Tuning and Evaluation of a JEV-Style Model for Tasks Involving Normative Knowledge}
\author{Qi Li\thanks{Contact: \href{mailto:liqi@czkj1010.com}{\texttt{liqi@czkj1010.com}}.} \quad Jianjun Xu \quad Pingtao Wei\\Beijing Caizhi Tech, Beijing, China}
\date{}
\begin{document}
\maketitle
\begin{abstract}
Tasks involving normative knowledge include concept interpretation, eligibility determination, procedure explanation, material requirements, and service-outcome assessment, all of which require rules and facts published, maintained, or recorded by authorized entities. Whether a candidate paragraph can answer a question therefore depends on applicability and answer completeness as well as topical relevance. We develop DKnown-Jev to classify the normative usability of candidate paragraphs into two classes. The 2B and 27B configurations are fine-tuned directly from the native Qwen3.5-2B and Qwen3.8-27B base models, respectively. The models combine question conditions, article metadata, and paragraph content, adopt the classes Answer Evidence (AE) and Non-answer Evidence (NE), and undergo domain adaptation using human-verified training data. Five configurations are evaluated on the same test set of 1,120 human-verified examples. DKnown-Jev 27B trained for one epoch achieves the highest accuracy of 98.75\% and macro-F1 of 98.60\%, exceeding Jev by 6.07 and 6.76 percentage points, respectively. DKnown-Jev 2B trained for one epoch achieves 92.86\% accuracy and 92.18\% macro-F1, the best aggregate results among the tested 2B configurations. Its gains over Jev are 0.18 and 0.33 percentage points on this test set, corresponding to only two additional correct predictions, which is insufficient to establish a stable advantage. The 27B one-epoch, 27B two-epoch, and 2B one-epoch configurations all exceed Jev on both metrics on this test set. Class-level analysis shows that the 27B two-epoch configuration has the highest AE recall, while the 27B one-epoch configuration has the highest NE recall. This study provides practical and empirical evidence for developing and applying candidate-scoring models to tasks involving normative knowledge.
\end{abstract}

\section{Introduction}
Tasks involving normative knowledge in this study refer to activities such as concept interpretation, eligibility determination, procedure explanation, material requirements, and service-outcome assessment that require rules and facts published, maintained, or recorded by authorized entities. Here, normative knowledge concerns domain rules, standards, and requirements; the tasks also draw on relevant facts. Their scope is not limited to legal force or administrative orders. An authorized entity is an actor responsible for rulemaking, information maintenance, or fact recording within the relevant domain, and is not necessarily an administrative agency. Institutional regulations, business standards, product manuals, policy documents, service catalogs, charging schedules, and related business records specify both which concepts hold under which conditions and which entities, regions, periods, and matters they apply to. Whether a paragraph can answer a question therefore cannot be determined solely by topical similarity.

Retrieval-augmented generation (RAG) combines retrieved external text with a generative model. Lewis et al.\ propose a representative framework combining parametric and non-parametric memory~\cite{lewis2020}. Dense retrieval and passage reranking can discover and select candidate text~\cite{karpukhin2020,nogueira2019}. These mechanisms effectively narrow the search space, but scores targeting semantic relevance alone are insufficient as the sole basis for judging paragraph usability under applicable rules and conditions. Two paragraphs may discuss the same matter using similar terms while applying to different regions, populations, periods, procedural stages, or rule versions. Conversely, a higher-level rule with broader scope that actually covers the question's conditions may provide a complete answer despite limited lexical overlap. Similarity-only ranking can mistake apparently relevant content for valid answer support.

Matching paragraphs containing normative knowledge and related facts therefore requires discrete relational judgments under applicability constraints, grounded in semantic understanding: whether a candidate paragraph directly answers the question or supplies a definite answer component, or at most provides background assistance. We express this distinction through two task-specific classes rather than by simply dividing continuous similarity scores into intervals.

Article metadata is an important part of this judgment. Paragraphs may omit regions, publication dates, issuing entities, or target populations that affect their suitability as answer support. DKnown-Jev jointly processes question conditions, available article metadata, and paragraph content to assess normative usability for a specific question. This concept encompasses rule- and fact-based judgments in institutional management, product services, and charging practices; the experiments use policy and public-service questions.

We adopt two task-specific classes: Answer Evidence (AE) and Non-answer Evidence (NE). The candidate-scoring architecture separately scores sequences combining the task input with each candidate definition and normalizes the scores to return class probabilities and predicted labels. Evaluation uses the AE/NE binary classification scheme.

Model performance must be evaluated through both aggregate classification quality and class-specific recognition. Training class distributions affect learning across classes. We investigate training data, domain adaptation, and classification evaluation.

Our work comprises paragraph-level training data with article metadata, domain adaptation with shared candidate scoring, and comparison of five configurations on the same evaluation examples with class-level error analysis.

\section{Related Work}
\subsection{Text Representations, Passage Retrieval, and Fine-Grained Classification}
Pretrained language models provide general representations for text classification and passage matching. BERT, proposed by Devlin et al., learns representations through bidirectional contextual modeling and adapts to classification, question answering, and natural language inference using task-specific output layers~\cite{devlin2019}. For passage retrieval, DPR, proposed by Karpukhin et al., uses dual encoders to represent questions and passages separately and retrieve candidates in vector space~\cite{karpukhin2020}. Nogueira and Cho study BERT-based passage reranking to assess query--passage relevance~\cite{nogueira2019}. These studies offer representative approaches to candidate retrieval and fine-grained matching, respectively.

ColBERT, proposed by Khattab and Zaharia, independently encodes queries and documents before fine-grained matching through late interaction~\cite{khattab2020}. It focuses on query--document relevance matching.

\subsection{Generative Classification, Prompt Learning, and Candidate Scoring}
T5, proposed by Raffel et al., unifies multiple natural language processing tasks in a text-to-text format, allowing classification through target-text generation~\cite{raffel2020}. PET, proposed by Schick and Sch\"utze, reformulates tasks as cloze-style patterns, connecting pretrained models to classification objectives through prompts and label words~\cite{schick2021}. These approaches show that class decisions can be expressed through text outputs or class-associated language; they differ in training objectives, output spaces, and inference mechanisms.

Nogueira et al.\ use a pretrained sequence-to-sequence model for document ranking, generating relevance label words and deriving ranking probabilities from their scores~\cite{nogueira2020}. Sun et al.\ investigate generative large language models as reranking agents that receive prompted candidate lists and output rankings~\cite{sun2023}. The former emphasizes label-word probabilities, whereas the latter emphasizes listwise ranking.

The expression of class meanings also affects task formulation. Yin et al.\ formulate zero-shot text classification as textual entailment, allowing linguistic class descriptions to participate in decisions~\cite{yin2019}.

\subsection{Domain Adaptation and Parameter-Efficient Fine-Tuning}
General pretrained capabilities may differ from target-domain requirements. Gururangan et al.\ study domain-adaptive and task-adaptive pretraining, showing that continued pretraining improves downstream performance across several domains and classification tasks in their experiments~\cite{gururangan2020}.

LoRA, proposed by Hu et al., freezes pretrained weights and represents parameter updates through trainable low-rank matrices, reducing the number of trainable parameters during fine-tuning~\cite{hu2022}.

\subsection{Evaluation Metrics}
The definition of macro-F1 must be explicit. Opitz and Burst distinguish the arithmetic mean of class-wise F1 scores from the harmonic mean of macro-averaged precision and recall. These formulas can produce different values and even different model rankings~\cite{opitz2019}.

\section{Application Task and Method}
\subsection{Normative Usability and Scope of Applicability}
For these tasks, normative usability denotes whether a candidate paragraph provides usable answer support under the question's applicability conditions. It is a relation between the paragraph and a specific question with its conditions, rather than an intrinsic paragraph attribute or a rating of a document's legal force. Classification considers applicability conditions such as population, region, time, subject matter, procedural stage, and rule version, as well as whether the paragraph provides sufficient answer information. The same paragraph may receive different labels for questions with different conditions.

Scope coverage must be supported by content or article metadata, not inferred solely from regional hierarchy, similar titles, or source names. Publication dates are not necessarily effective dates, and issuing-entity information does not automatically establish current validity. The model can judge only from visible information; missing necessary conditions or conflicts should not be resolved by treating topical similarity as definitive answer support.

\subsection{Binary Classification Scheme}
The input comprises the user's question and its conditions $q$, article metadata $a$, and a candidate paragraph $p$, denoted $x=(q,a,p)$. The class set is $\{\mathrm{AE},\mathrm{NE}\}$. Metadata supplies context such as region, time, and title that the paragraph itself may omit.

\begin{table}[htbp]
\centering\small
\caption{Definitions of the two task-specific classes.}\label{tab:classes}
\begin{tabularx}{\linewidth}{@{}l l Y@{}}
\toprule Abbreviation & Name & Definition\\\midrule
AE & Answer Evidence & The paragraph directly answers the question or supplies a definite component of the answer.\\
NE & Non-answer Evidence & The paragraph merely assists interpretation of the question or has no substantive relation to it.\\
\bottomrule
\end{tabularx}
\end{table}

Direct answers include both cases where the scope matches the question and cases where a broader scope covers the question's conditions. Supplying an answer component includes paragraphs that answer only part of the question's requirements. In either case, the paragraph carries usable answer content and belongs to AE. Paragraphs that merely assist interpretation or are unrelated to the question belong to NE; lexical overlap alone cannot determine the class. Judgments use the content and metadata actually supplied without inventing missing conditions. Table~\ref{tab:classes} defines AE and NE.

\subsection{Data Structure and Field Contents}
Each record comprises task input \texttt{state}, task instruction \texttt{question}, candidate definitions \texttt{options}, and target distribution \texttt{target}. The \texttt{state} contains \texttt{query} and \texttt{article}: \texttt{query} records question conditions such as the user's question and service region; \texttt{article} records metadata such as title, publication date, and region, and stores the candidate paragraph's title and body in a \texttt{paragraph} object. The following schematic translates the descriptive Chinese field names into English:
\begin{lstlisting}
{
  "state": {
    "query": {"user_question": "...", "service_region": "..."},
    "article": {
      "article_title": "...",
      "publication_date": "...",
      "regions": ["..."],
      "paragraph": {"title": "...", "content": "..."}
    }
  }
}
\end{lstlisting}
The \texttt{question} field describes the paragraph-evidence classification task; \texttt{options} provides the AE and NE candidate definitions; and \texttt{target} records the target probability distribution in candidate order. Together, these fields describe the question and paragraph to be assessed, the classification requirements, and the supervision.

\subsection{Shared Candidate Scoring and Training Objective}\label{sec:scoring}
We adopt a shared candidate-scoring architecture and training mechanism, organizing task inputs, defining class semantics, and performing domain fine-tuning for tasks involving normative knowledge. The architecture retains the pretrained language-model backbone and attaches a shared scalar scoring head to its output representations. Instead of using the original vocabulary output head to generate classification answers token by token, scoring directly computes the match between each candidate and the task input.

For class $k$, task input $x$, task instruction $u$, and candidate definition $d_k$ form a prompt $T(x,u,d_k)$. The backbone supplies the representation $h_k$ at the last valid input position. All candidates share the same backbone and scoring head, each yielding a scalar score $s_k$; softmax then produces class probabilities $p_k$. For the binary task:
\begin{equation}
s_k=w^\top h_k+b,\qquad p_k=\frac{\exp(s_k)}{\sum_{j=1}^{2}\exp(s_j)}.
\end{equation}
Here, $w$ and $b$ are the shared head's weight and bias. The weight $w$ is initialized with the difference between the weight vectors for Yes and No in the base model's original vocabulary output head, and $b$ is initialized to zero. This allows candidate discrimination to start from the pretrained model's existing affirmative--negative output difference. Candidate sequences can be processed in a batch, but each separately contains the task input and its candidate definition.

Parameter updates use LoRA~\cite{hu2022}: the original backbone parameters are frozen, and trainable low-rank adapters are introduced into the attention-related projection layers available in the model. Available layers are matched among target modules such as query, key, value, and output projections. Training updates only these LoRA parameters and the shared scoring head, leaving the original backbone parameters frozen. LoRA adapts representations of the task input and candidate definitions, while the head learns to convert them into candidate scores; both are jointly optimized by the same loss.

Training combines cross-entropy with a Brier term, which measures squared error between predicted probabilities and the target distribution~\cite{brier1950}. The Brier term has weight 0.1. Let $y_k$ be the target probability and $p_k$ the predicted probability for class $k$. The loss is
\begin{equation}
L=-\sum_{k=1}^{2}y_k\log p_k+0.1\sum_{k=1}^{2}(p_k-y_k)^2.
\end{equation}
Cross-entropy encourages higher relative probability for the correct class; the Brier term constrains squared probability error with weight 0.1. Gradients propagate through the shared scoring head to the LoRA parameters, jointly adapting candidate representations and the scoring rule to task supervision. This objective directly optimizes the candidate class distribution and does not use a token-by-token answer-generation loss.

Domain adaptation uses single-label supervision over AE and NE. The 2B configuration is fine-tuned directly from the native Qwen3.5-2B base model, and the 27B configuration from the native Qwen3.8-27B base model. Section~\ref{sec:data} describes the training data, and Table~\ref{tab:models} lists model scales and training epochs.

\section{Training Data}\label{sec:data}
\subsection{Example Organization and Input Fields}
The training set contains 26,064 human-verified records, each pairing a question with a candidate paragraph. Inputs comprise question conditions, article metadata, and paragraph content. Question fields include the user's question and available service region; article fields include title, region, and publication date; paragraph objects contain a title and body. Article metadata and paragraph text jointly provide the classification context.

Training records use the \texttt{choice} task format with \texttt{state}, \texttt{question}, \texttt{options}, and \texttt{target} fields. These represent task input, the classification task, AE and NE candidate meanings, and the two-class target probability distribution, respectively. The model interprets class meanings through candidate definitions rather than through the abbreviations themselves.

\subsection{Training Set Size and Class Distribution}
\begin{table}[htbp]
\centering
\caption{Training class distribution.}\label{tab:training}
\begin{tabular}{lrr}
\toprule Class & Count & Proportion (\%)\\\midrule
AE & 11,788 & 45.23\\
NE & 14,276 & 54.77\\
Total & 26,064 & 100.00\\\bottomrule
\end{tabular}
\end{table}
NE accounts for 54.77\% of the training set and AE for 45.23\%, a relatively balanced distribution. Evaluation nevertheless reports macro-F1 and class recall alongside accuracy to prevent aggregate metrics from obscuring class-specific recognition differences.

\subsection{Information Boundaries in Training and Inference}
Target labels contribute only to the training loss and are excluded from task inputs. Inference supplies only the question, article metadata, paragraph, and candidate definitions. Evaluation follows the binary definitions in Table~\ref{tab:classes}.

\section{Experimental Setup}
\subsection{Test Set and Model Sources}
We evaluate on a fixed set of 1,120 human-verified examples: 378 AE and 742 NE. Aggregate metrics use these class proportions. Each test example has an AE or NE ground-truth label, and predictions from the five configurations are aligned by example identifier.

The comparison includes commercial Jev and four domain-adapted models. A1 and A2 are fine-tuned directly from the native \texttt{Qwen/Qwen3.5-2B} base~\cite{qwen2b} for one and two domain epochs, respectively. B1 and B2 are fine-tuned directly from the native Qwen3.8-27B base~\cite{qwen27b} for one and two domain epochs, respectively. Epoch counts refer only to the domain-training stage in this study. All four configurations are deployed with the architecture in Section~\ref{sec:scoring} and evaluated independently.

\begin{table}[htbp]
\centering\small
\caption{Model initialization and domain-training configurations.}\label{tab:models}
\begin{tabularx}{\linewidth}{@{}l Y c@{}}
\toprule Model or service & Initialization source & \shortstack{Domain-training\\epochs}\\\midrule
Jev & Not publicly confirmed & Not publicly confirmed\\
A1: DKnown-Jev 2B & Qwen3.5-2B & 1\\
A2: DKnown-Jev 2B & Qwen3.5-2B & 2\\
B1: DKnown-Jev 27B & Qwen3.8-27B & 1\\
B2: DKnown-Jev 27B & Qwen3.8-27B & 2\\\bottomrule
\end{tabularx}
\end{table}
Table~\ref{tab:models} records deployment initialization sources, model scales, and training epochs as confirmed by the researchers. A1, A2, B1, and B2 identify the four DKnown-Jev configurations below.

\subsection{Evaluation Metrics and Comparison Procedure}
Accuracy is the number of correctly predicted examples divided by the test set size. Macro-F1 is the arithmetic mean of the two class-wise F1 scores, each computed from that class's precision and recall. Class recall is also reported, and confusion matrices inform the analysis of error directions.

Commercial Jev and all four DKnown-Jev configurations complete predictions for the same 1,120 examples, yielding 5,600 predictions. Evaluation follows the AE/NE binary classification scheme, comparing each configuration's predictions with ground-truth labels to calculate accuracy, macro-F1, and class recall consistently. Every example has predictions from all five configurations, with no duplicate or missing identifiers.

\section{Experimental Results}
\subsection{Aggregate Classification Performance}
\begin{table}[htbp]
\centering
\caption{Binary classification performance on 1,120 test examples.}\label{tab:results}
\begin{tabular}{lrrr}
\toprule Model or service & Correct & Accuracy (\%) & Macro-F1 (\%)\\\midrule
Jev & 1,038 & 92.68 & 91.85\\
A1 (2B, 1 epoch) & 1,040 & 92.86 & 92.18\\
A2 (2B, 2 epochs) & 1,020 & 91.07 & 90.19\\
B1 (27B, 1 epoch) & 1,106 & 98.75 & 98.60\\
B2 (27B, 2 epochs) & 1,055 & 94.20 & 93.74\\\bottomrule
\end{tabular}
\par\smallskip
{\footnotesize A1, A2, B1, and B2 are DKnown-Jev configurations; see Table~\ref{tab:models}.}
\end{table}
B1 achieves the highest accuracy and macro-F1, at 98.75\% and 98.60\%, exceeding Jev by 6.07 and 6.76 percentage points, respectively. A1 achieves 92.86\% and 92.18\%, exceeding Jev by 0.18 and 0.33 points on this test set, but only makes two additional correct predictions, which is insufficient to establish a stable advantage. B2 achieves 94.20\% and 93.74\%, exceeding Jev by 1.52 and 1.89 points. On this test set, B1, B2, and A1 exceed Jev on both metrics, whereas A2 trails it by 1.61 and 1.66 points.

At the same model scale, B1 exceeds B2 by 4.55 accuracy points and 4.86 macro-F1 points; A1 exceeds A2 by 1.79 and 1.99 points. For these configurations and this test set, one epoch yields higher values on both aggregate metrics than two epochs at each scale. This observation cannot be directly generalized into a claim that increasing training epochs reduces performance.

At matching domain-training epochs, B1 exceeds A1 on both metrics by 5.89 and 6.42 points, and B2 exceeds A2 by 3.13 and 3.55 points. The 2B and 27B configurations differ in both parameter scale and model family, so these comparisons concern complete configurations. Across configurations, B2 exceeds A1 by 1.34 accuracy points and 1.56 macro-F1 points.

\subsection{Class Recall and Error Patterns}
\begin{table}[htbp]
\centering\small
\caption{Class recall (\%); parentheses indicate class counts.}\label{tab:recall}
\begin{tabular}{lrr}
\toprule Model or service & AE (378) & NE (742)\\\midrule
Jev & 89.95 & 94.07\\
A1 (2B, 1 epoch) & 93.92 & 92.32\\
A2 (2B, 2 epochs) & 90.48 & 91.37\\
B1 (27B, 1 epoch) & 97.88 & 99.19\\
B2 (27B, 2 epochs) & 99.47 & 91.51\\\bottomrule
\end{tabular}
\par\smallskip
{\footnotesize A1, A2, B1, and B2 are DKnown-Jev configurations; see Table~\ref{tab:models}.}
\end{table}
The highest class recalls occur in different configurations: B2 achieves 99.47\% AE recall, while B1 achieves 99.19\% NE recall. No model achieves the highest recall for both classes.

B1, which has the highest accuracy and macro-F1, misclassifies 8 of 378 AE examples as NE and 6 of 742 NE examples as AE, with few errors in either direction. B2 misses only 2 AE examples but classifies 63 NE examples as AE. A1 incorrectly accepts 57 NE examples and misses 23 AE examples; the corresponding counts for A2 are 64 and 36. Jev misses 38 AE examples and incorrectly accepts 44 NE examples.

\section{Discussion and Limitations}
This study uses candidate scoring to distinguish usable answer support from background material under the question's applicability conditions. Evaluation measures correct predictions under the specified classification criteria. Overall accuracy and class recall reflect different aspects of performance and should be analyzed together.

The decision boundary remains challenging: when a paragraph provides content resembling a definition, rule, or procedural step, determining whether it constitutes a definite answer component requires joint assessment of applicability and answer completeness rather than topical relevance alone.

Applications differ in their requirements for the two error types. Those requiring high answer coverage should prioritize AE recall; those needing to avoid unsupported answer claims should prioritize NE recall.

The test set contains 378 AE and 742 NE examples. B2 has the highest AE recall, but its NE recall is 91.51\%, below B1's 99.19\%. Both of B1's class recalls exceed 97\%, illustrating why deployment selection should not rely solely on a single ranking. Classifying NE as AE can cause downstream generation to treat reference or unrelated material as answer support; missing AE can omit usable material.

\section{Conclusion}
Motivated by the distinction between semantic relevance and valid answer support in tasks involving normative knowledge, this study develops DKnown-Jev using candidate scoring. It assesses normative usability from question conditions, article metadata, and paragraph content, and evaluates services using binary classification on the same 1,120 human-verified test examples.

Among the five configurations, DKnown-Jev 27B trained for one epoch achieves the highest accuracy of 98.75\% and macro-F1 of 98.60\%, exceeding Jev by 6.07 and 6.76 percentage points, respectively. DKnown-Jev 2B trained for one epoch achieves 92.86\% and 92.18\%, the best aggregate results among the tested 2B configurations. Its gains over Jev are 0.18 and 0.33 points on this test set, corresponding to only two additional correct predictions, which is insufficient to establish a stable advantage. The 27B one-epoch, 27B two-epoch, and 2B one-epoch configurations all exceed Jev on both metrics on this test set.

On this test set, B2 misses the fewest AE examples, while B1 incorrectly accepts the fewest NE examples and achieves the highest accuracy and macro-F1.

\FloatBarrier


\begin{thebibliography}{16}
\bibitem{lewis2020}
P. Lewis, E. Perez, A. Piktus, et al.
Retrieval-Augmented Generation for Knowledge-Intensive NLP Tasks.
arXiv:2005.11401, 2020.
\url{https://arxiv.org/abs/2005.11401}.

\bibitem{karpukhin2020}
V. Karpukhin, B. Oguz, A. Piktus, et al.
Dense Passage Retrieval for Open-Domain Question Answering.
In \emph{EMNLP}, pp. 6769--6781, 2020.
\url{https://doi.org/10.18653/v1/2020.emnlp-main.550}.
Publication page: \url{https://aclanthology.org/2020.emnlp-main.550/}.

\bibitem{nogueira2019}
R. Nogueira and K. Cho.
Passage Re-ranking with BERT.
arXiv:1901.04085, 2019.
\url{https://arxiv.org/abs/1901.04085}.

\bibitem{devlin2019}
J. Devlin, M.-W. Chang, K. Lee, and K. Toutanova.
BERT: Pre-training of Deep Bidirectional Transformers for Language Understanding.
In \emph{NAACL-HLT}, pp. 4171--4186, 2019.
\url{https://doi.org/10.18653/v1/N19-1423}.
Publication page: \url{https://aclanthology.org/N19-1423/}.

\bibitem{khattab2020}
O. Khattab and M. Zaharia.
ColBERT: Efficient and Effective Passage Search via Contextualized Late Interaction over BERT.
arXiv:2004.12832, 2020.
\url{https://arxiv.org/abs/2004.12832}.

\bibitem{raffel2020}
C. Raffel, N. Shazeer, A. Roberts, et al.
Exploring the Limits of Transfer Learning with a Unified Text-to-Text Transformer.
\emph{Journal of Machine Learning Research}, 21(140):1--67, 2020.
\url{https://www.jmlr.org/papers/v21/20-074.html}.

\bibitem{schick2021}
T. Schick and H. Sch\"utze.
Exploiting Cloze-Questions for Few-Shot Text Classification and Natural Language Inference.
In \emph{EACL}, pp. 255--269, 2021.
\url{https://doi.org/10.18653/v1/2021.eacl-main.20}.
Publication page: \url{https://aclanthology.org/2021.eacl-main.20/}.

\bibitem{nogueira2020}
R. Nogueira, Z. Jiang, and J. Lin.
Document Ranking with a Pretrained Sequence-to-Sequence Model.
arXiv:2003.06713, 2020.
\url{https://arxiv.org/abs/2003.06713}.

\bibitem{sun2023}
W. Sun, L. Yan, X. Ma, et al.
Is ChatGPT Good at Search? Investigating Large Language Models as Re-Ranking Agents.
arXiv:2304.09542, 2023.
\url{https://arxiv.org/abs/2304.09542}.

\bibitem{yin2019}
W. Yin, J. Hay, and D. Roth.
Benchmarking Zero-shot Text Classification: Datasets, Evaluation and Entailment Approach.
arXiv:1909.00161, 2019.
\url{https://arxiv.org/abs/1909.00161}.

\bibitem{gururangan2020}
S. Gururangan, A. Marasovi\'{c}, S. Swayamdipta, et al.
Don't Stop Pretraining: Adapt Language Models to Domains and Tasks.
In \emph{ACL}, pp. 8342--8360, 2020.
\url{https://doi.org/10.18653/v1/2020.acl-main.740}.
Publication page: \url{https://aclanthology.org/2020.acl-main.740/}.

\bibitem{hu2022}
E. J. Hu, Y. Shen, P. Wallis, et al.
LoRA: Low-Rank Adaptation of Large Language Models.
In \emph{ICLR}, 2022.
\url{https://openreview.net/forum?id=nZeVKeeFYf9}.
Author preprint: \url{https://arxiv.org/abs/2106.09685}.

\bibitem{opitz2019}
J. Opitz and S. Burst.
Macro F1 and Macro F1.
arXiv:1911.03347, 2019.
\url{https://arxiv.org/abs/1911.03347}.

\bibitem{brier1950}
G. W. Brier.
Verification of Forecasts Expressed in Terms of Probability.
\emph{Monthly Weather Review}, 78(1):1--3, 1950.
\url{https://doi.org/10.1175/1520-0493(1950)078<0001:VOFEIT>2.0.CO;2}.

\bibitem{qwen2b}
Qwen Team.
Qwen3.5-2B Model Card.
\url{https://modelscope.cn/models/Qwen/Qwen3.5-2B}.
Accessed October 13, 2026.

\bibitem{qwen27b}
Qwen Team.
Qwen3.8-27B Model Card.
\url{https://modelscope.cn/models/Qwen/Qwen3.8-27B}.
Accessed October 13, 2026.
\end{thebibliography}
\end{document}